\newcommand{\PaperAuthorName}{Bingxuan Xie}
\newcommand{\PaperAffiliation}{South China Normal University}
\newcommand{\PaperEmail}{20232731078@m.scnu.edu.cn}

\documentclass{article}

\PassOptionsToPackage{numbers,sort&compress}{natbib}
\usepackage[preprint]{neurips_2019}
\makeatletter
\renewcommand{\@notice}{}
\makeatother
\usepackage[utf8]{inputenc}
\usepackage[T1]{fontenc}
\usepackage{amsmath}
\usepackage{newtxtext,newtxmath}
\usepackage{graphicx}
\usepackage{booktabs}
\usepackage{algorithm}
\usepackage{algpseudocode}
\usepackage{microtype}
\usepackage{enumitem}
\usepackage{flafter}
\usepackage{placeins}
\usepackage{url}
\usepackage[hidelinks]{hyperref}
\usepackage{doi}

\newcommand{\dfit}{\mathcal{D}_{\mathrm{fit}}}
\newcommand{\dmeta}{\mathcal{D}_{\mathrm{meta}}}
\newcommand{\dtest}{\mathcal{D}_{\mathrm{test}}}
\newcommand{\stopgrad}{\operatorname{sg}}
\newcommand{\clip}{\operatorname{clip}}

\title{Dynamic Influence-Weighted Distillation for\\
Single-IMU Activity Recognition}

\author{%
  \PaperAuthorName\\
  \normalfont \PaperAffiliation\\
  \normalfont \texttt{\PaperEmail}%
}

\date{}

\hypersetup{
  pdftitle={Dynamic Influence-Weighted Distillation for Single-IMU Activity Recognition},
  pdfauthor={\PaperAuthorName},
  pdfsubject={Wearable activity recognition with training-only multi-position IMU signals},
  pdfkeywords={wearable activity recognition, inertial sensing, privileged information, knowledge distillation, dynamic weighting}
}

\begin{document}
\raggedbottom
\maketitle

\begin{abstract}
Inertial sensors at multiple body locations can improve activity recognition, but requiring every sensor at inference increases the deployment burden. We study whether four synchronized IMUs available during training can improve a student that uses only the right-arm IMU during fitting and inference. A frozen four-IMU teacher provides logit and feature targets. Fixed-weight knowledge distillation applies each target with the same strength to every fitting sample, although the student may not benefit equally from them. We introduce dynamic influence weighting (DIW), which tests a one-step candidate update on separate fold-internal training participants. DIW then assigns separate sample-wise gates to the logit and feature losses. On WEAR, we evaluate 19 labels and 68,298 complete windows from 22 participants using subject-disjoint five-fold cross-validation. Pooled out-of-fold macro-F1 is 0.561820 for Supervised and 0.571623 for Fixed-weight KD. DIW reaches 0.638451, gains of 7.66 and 6.68 percentage points, respectively. It exceeds Supervised for 18 of 19 labels and 21 of 22 held-out participants. All three routes retain the same 80,915-parameter right-arm student at inference. Under this protocol, DIW converts training-only multi-position information into a stronger single-IMU model without changing deployed sensing or the student forward graph.
\end{abstract}

\section{Introduction}
\label{sec:introduction}

Human activity recognition from body-worn inertial streams has progressed from engineered pipelines to deep temporal models~\cite{bulling2014tutorial,lara2013survey,hammerla2016deep,ordonez2016deep,gu2021survey}. A sensor at one body location captures only part of an action. Lower-body sensors emphasize periodic leg motion, whereas an arm sensor directly captures many upper-body exercises. Combining locations therefore provides broader motion coverage.

The same coverage makes deployment harder. Every additional device must be worn, synchronized, powered, and maintained. A controlled data collection can tolerate this burden more readily than routine use. Prior work has therefore used additional training sensors, cross-location representation learning, and virtual sensor fusion while retaining one target sensor at inference~\cite{lago2021additional,rey2022learning,nguyen2024virtual}. We adopt this setting with four synchronized training IMUs and a final classifier that receives only the right-arm IMU.

Learning using privileged information formalizes access to observations that are unavailable to the deployed predictor~\cite{vapnik2009lupi}. Generalized distillation applies this idea through teacher--student learning~\cite{lopezpaz2016generalized}. A four-IMU teacher can provide softened class targets~\cite{hinton2015distilling} and intermediate feature targets~\cite{romero2015fitnets} to a right-arm student. However, a more accurate teacher does not guarantee useful guidance for every student update. The teacher may rely on leg or left-arm motion that the right-arm student cannot observe. At any given step, Fixed-weight KD nevertheless applies the same logit and feature coefficients to every fitting sample.

We introduce \emph{dynamic influence weighting} (DIW) to decide how strongly each teacher target should affect the current student. DIW first constructs a detached one-step candidate update and evaluates it on a fold-internal meta set. It then probes each fitting sample's logit and feature losses along the resulting meta-feedback direction. Resolved positive alignment is mapped separately to two gates in $[0,1]$. The teacher, meta set, and gate-estimation operations are used only during training.

We compare a supervised right-arm student, the same student with Fixed-weight KD, and the same student with DIW on WEAR~\cite{bock2024wear}. Under subject-disjoint five-fold evaluation, pooled out-of-fold (OOF) macro-F1 is 0.561820, 0.571623, and 0.638451, respectively. DIW exceeds Supervised for 18 of 19 labels and 21 of 22 held-out participants. Fold-0 gate diagnostics and targeted ablations examine how the learned selection changes across knowledge components, training stages, and activity classes.

Our contributions are threefold:
\begin{itemize}[leftmargin=1.35em,itemsep=0.25em,topsep=0.35em]
  \item We define a controlled four-IMU teacher and right-arm student protocol with identical single-IMU inference across all deployable routes.
  \item We develop DIW, which combines a detached look-ahead assessment with sample-wise gates estimated independently for logit and feature distillation.
  \item We evaluate the three routes on common OOF predictions and relate aggregate performance to label-level, participant-level, gate-diagnostic, and ablation evidence.
\end{itemize}

\section{Related Work}
\label{sec:related_work}

\subsection{Rich sensing during training and restricted sensing at inference}

Learning using privileged information permits auxiliary observations during training while requiring the final predictor to operate without them~\cite{vapnik2009lupi}. Distillation provides a general mechanism for transferring such training-only information into a model defined on the deployment input~\cite{lopezpaz2016generalized}. Related work has transferred side information through modality hallucination and privileged recurrent representations~\cite{hoffman2016hallucination,shi2017privileged}. In wearable sensing, the distinction is especially useful because a curated data collection can temporarily instrument more body locations than a person is expected to wear in routine use.

Several activity-recognition studies instantiate this training/deployment asymmetry directly. Lago \emph{et al.} use additional body-worn sensors during training to improve recognition from one retained sensor~\cite{lago2021additional}. Fortes Rey \emph{et al.} align representations across sensor locations and evaluate the target location alone~\cite{rey2022learning}. Virtual Fusion jointly exploits synchronized training sensors through contrastive learning while supporting a single-sensor inference branch~\cite{nguyen2024virtual}. These approaches show that temporary sensing can inform a predictor whose eventual input is more restricted.

Related systems extend the same idea to heterogeneous modalities or explicit teacher--student learning. MESEN uses multimodal data during model design to support unimodal activity recognition with few labels~\cite{xu2023mesen}. TSAK transfers semantic representations from a multi-position, multimodal teacher to a smaller single-hand model~\cite{bello2025tsak}. Sensor-to-Sensor Procedural Co-learning aligns a sensor-rich teacher and a sensor-limited student through input adaptation and multi-level feature objectives~\cite{xie2026sensor}. These studies use different forms of training-only information, but all aim to improve a model with fewer inputs at deployment.

\subsection{Distillation and validation-guided weighting}

Model compression established the broader premise that a compact predictor can learn from a stronger model~\cite{bucilua2006compression}. Classical KD trains a student against the softened output distribution of a teacher~\cite{hinton2015distilling}, while feature-based methods align internal representations or attention patterns~\cite{romero2015fitnets,zagoruyko2017attention}. Surveys organize these targets within the wider distillation literature~\cite{gou2021survey}. Combining output and feature objectives is common when the teacher contains information not directly exposed by the student's input, but constant loss coefficients treat each fitting sample identically.

Validation-guided reweighting provides one way to adapt training contributions. Learning to Reweight Examples derives current-batch weights from agreement between training and clean-validation gradient directions~\cite{ren2018reweight}. Meta-Weight-Net instead learns a loss-to-weight mapping from a meta set~\cite{shu2019metaweightnet}. Dynamic-loss teaching adapts a learned objective to the student's state and training stage~\cite{wu2018dynamicloss}. Classical influence functions estimate how a small change in training weight affects a model prediction~\cite{koh2017influence}. These studies use feedback beyond the instantaneous training loss to decide how strongly a sample or objective should contribute.

Adaptive weighting has also been studied inside distillation. Multi-teacher KD can learn separate combinations of output- and feature-level information~\cite{zhang2023adaptive}. LGTM defines a validation-based distillation influence for individual training examples and uses a finite-difference approximation to shape teacher learning~\cite{ren2023tailoring}. Across these approaches, adaptive distillation can act on different objects, including teacher combinations, training examples, and the teacher's own update.

\section{Method}
\label{sec:method}

\subsection{Problem setting}
\label{sec:problem_setting}

Let $\mathcal{P}=\{\mathrm{RA},\mathrm{RL},\mathrm{LL},\mathrm{LA}\}$ denote the right-arm, right-leg, left-leg, and left-arm sensor positions. A synchronized training example is $(X_i^{\mathcal{P}},y_i)$, where $X_i^{\mathcal{P}}=\{x_i^p:p\in\mathcal{P}\}$ and $y_i\in\{1,\ldots,K\}$ with $K=19$. The four-IMU teacher $T_{\phi}$ receives $X_i^{\mathcal{P}}$. The right-arm student $S_{\theta}$ receives only $x_i^{\mathrm{RA}}$ during fitting and inference. Thus, the teacher's non-right-arm inputs are privileged training information rather than missing inputs that the student must reconstruct.

Within each outer cross-validation fold, the outer-training participants are divided into a fitting set $\dfit$ and a fold-internal meta set $\dmeta$. Three participants form $\dmeta$, and the remaining outer-training participants form $\dfit$. The fitting set supplies normalization statistics, class weights, teacher training data, and direct student updates. DIW uses the meta set only to evaluate candidate distillation updates. The participant sets in $\dfit$, $\dmeta$, and the outer test fold $\dtest$ are pairwise disjoint. The outer test fold does not affect training, gate estimation, epoch count, or checkpoint selection.

\begin{figure}[tbp]
  \centering
  \includegraphics[width=\linewidth]{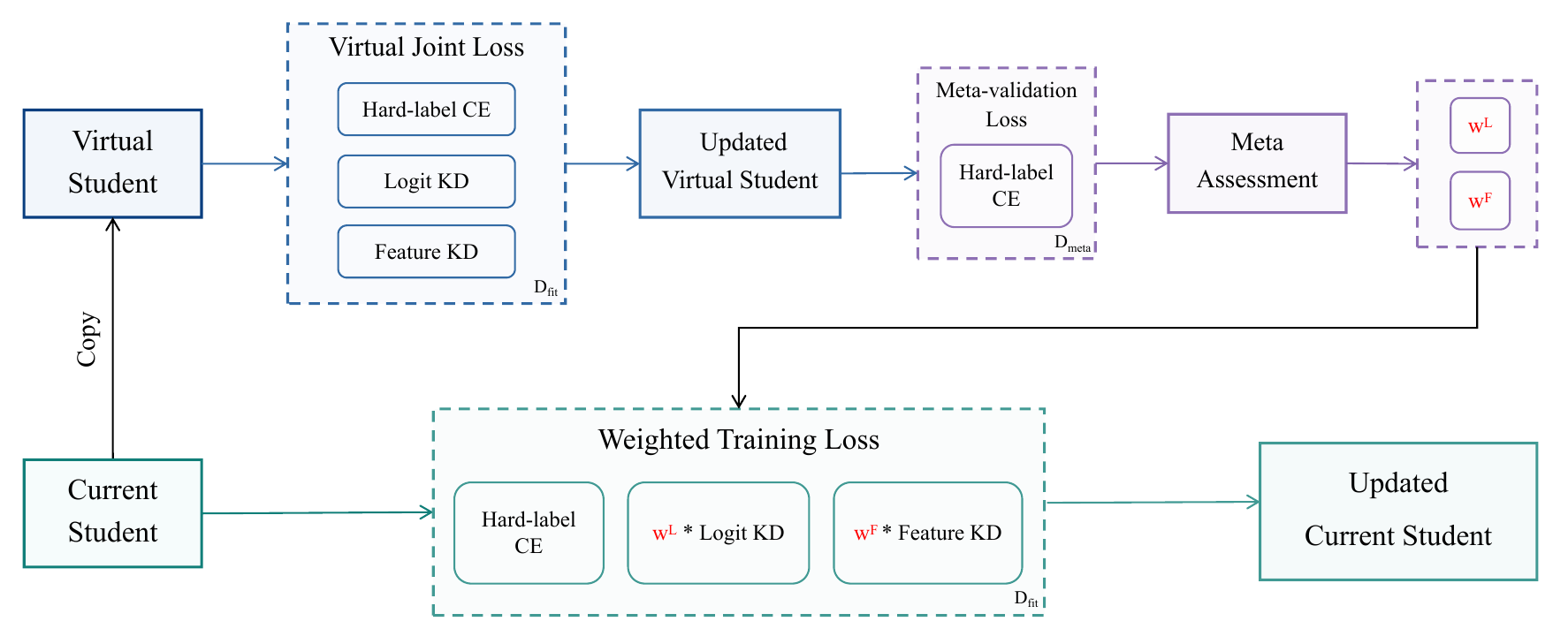}
  \caption{DIW during student training. A detached one-step update is evaluated on a fold-internal meta batch, and finite-difference probes yield separate sample-wise gates for logit and feature distillation. Only the right-arm student is retained for inference.}
  \label{fig:framework}
\end{figure}

\subsection{Inputs and networks}
\label{sec:networks}

Each position supplies a 50-sample, three-axis acceleration window. For the first-difference channels, we prepend the first sample so that the initial difference is exactly zero. We also append the signal magnitude vector (SMV)
\begin{equation}
  \operatorname{SMV}_t=\sqrt{a_{x,t}^{2}+a_{y,t}^{2}+a_{z,t}^{2}}.
  \label{eq:magnitude}
\end{equation}
The resulting channel order is $[a_x,a_y,a_z,\Delta a_x,\Delta a_y,\Delta a_z,\operatorname{SMV}]$, giving a $50\times7$ tensor per position. The difference channels are raw first temporal differences, not derivatives normalized by the sampling interval. Means and population standard deviations are estimated separately for each position and channel from $\dfit$; standard deviations are floored at $10^{-6}$.

The right-arm student is a compact temporal convolutional model~\cite{bai2018tcn}. It uses two residual one-dimensional convolutional blocks~\cite{he2016resnet} with 64 and 96 channels, kernel size 5, dilations 1 and 2, batch normalization, GELU activations, and dropout of 0.15. Concatenated temporal mean and maximum pooling produces a 192-dimensional representation $h_i^S$, followed by a $192\!\rightarrow\!192$ projection and a 19-class linear head. The student has 80,915 trainable parameters.

The four-IMU teacher applies one shared temporal encoder to the four positions in the fixed order RA, RL, LL, LA. Their four 192-dimensional representations are concatenated and passed through a learned $768\!\rightarrow\!192$ fusion projection, producing $h_i^T$, before the 19-class head. The resulting shared-encoder fusion teacher has 191,507 parameters. Teacher optimization has two stages. First, the shared encoder and classifier are trained for 32 epochs using the mean of four position-specific hard-label losses. The fusion stage loads this checkpoint and trains the complete teacher for another 32 epochs with a fresh optimizer and scheduler. The final logits $z_i^T$ and fused features $h_i^T$ on $\dfit$ are cached in FP32, and $\phi$ remains frozen throughout student training.

\subsection{Supervised and fixed-weight distillation}
\label{sec:fixed_objective}

Let $N_c$ be the number of fitting samples in class $c$, $N=|\dfit|$, and
$K=19$. The supervised term uses clipped inverse-frequency weighting,
\begin{equation}
  \alpha_c=\clip\!\left(\frac{N}{K N_c},1,4\right),
  \qquad
  \ell_i^{\mathrm{CE}}(\theta)
  =-\alpha_{y_i}\log p_S\!\left(y_i\mid x_i^{\mathrm{RA}};\theta\right).
  \label{eq:hard_loss}
\end{equation}
For the frozen teacher, let $z_i^T$ and $h_i^T$ denote its logits and fused
feature, and let $z_i^S(\theta)$ and $h_i^S(\theta)$ be the corresponding
student quantities. With $T_{\mathrm{kd}}=2$ and $d=192$, the two
sample-wise distillation components are
\begin{align}
  q_i&=\operatorname{softmax}\!\left(z_i^T/T_{\mathrm{kd}}\right),
  &p_i(\theta)&=\operatorname{softmax}\!\left(z_i^S(\theta)/T_{\mathrm{kd}}\right),
  \notag\\
  \ell_i^L(\theta)
    &=T_{\mathrm{kd}}^2D_{\mathrm{KL}}\!\left(q_i\,\|\,p_i(\theta)\right),
  &\ell_i^F(\theta)
    &=\frac{1}{d}\left\|h_i^S(\theta)-h_i^T\right\|_2^2.
  \label{eq:kd_components}
\end{align}
Teacher targets are detached. Superscripts $L$ and $F$ identify logit and
feature distillation, respectively, and remain separate throughout DIW.

Both distillation routes use base coefficients $\lambda_L=0.2$ and
$\lambda_F=0.05$. Their common ramp $\rho_e$ is zero through epoch 5,
equals $(e-5)/4$ for epochs 6--8, and equals one from epoch 9 onward. For a
fitting mini-batch $\mathcal{B}_f$ of size $B$, fixed-weight KD minimizes
\begin{equation}
  \mathcal{L}_{\mathrm{fixed}}(\theta;\mathcal{B}_f)
  =\frac{1}{B}\sum_{i\in\mathcal{B}_f}
   \left[
     \ell_i^{\mathrm{CE}}(\theta)
     +\rho_e\!\sum_{k\in\{L,F\}}\lambda_k\ell_i^k(\theta)
   \right].
  \label{eq:fixed_kd}
\end{equation}
At a given step, every fitting sample receives the same coefficient
$\rho_e\lambda_k$ for component $k$. The two components retain distinct base
coefficients, while the common ramp changes with epoch.

\subsection{Dynamic influence weighting}
\label{sec:diw}

DIW decides separately how much the logit and feature targets for each fitting
sample should contribute. For sample $i$ and component $k$, it asks whether
decreasing $\ell_i^k$ aligns locally with a direction that lowers hard-label
loss on fold-internal meta participants. The estimate is recomputed at every
fitting step and independently for $k=L$ and $k=F$. We call $\theta_t$ the
persistent student because these parameters continue into the next training
step.

\paragraph{Detached look-ahead and meta feedback.}
DIW first evaluates equation~\eqref{eq:fixed_kd} functionally in training mode with a
clone of the current batch-normalization buffers. Let $\mathbf{s}_t$ denote
the current AdamW state, let $C=5$ be the clipping norm, and let
$\mathsf{A}_t$ denote one AdamW state transition given the clipped gradient.
The look-ahead parameters and meta-feedback direction are
\begin{align}
  \mathbf{q}_t
    &=\operatorname{clip}_{C}\!\left(
        \nabla_{\theta_t}\mathcal{L}_{\mathrm{fixed}}
        (\theta_t;\mathcal{B}_f)\right),
  &\widetilde{\theta}_t
    &=\stopgrad\!\left[
        \mathsf{A}_t(\theta_t,\mathbf{q}_t;\mathbf{s}_t)\right],
  \notag\\
  \mathcal{L}_{\mathrm{meta}}(\widetilde{\theta}_t;\mathcal{B}_m)
    &=\frac{1}{|\mathcal{B}_m|}
      \sum_{r\in\mathcal{B}_m}\ell_r^{\mathrm{CE}}(\widetilde{\theta}_t),
  &\widehat{\mathbf{g}}_m
    &=\frac{\nabla_{\widetilde{\theta}_t}\mathcal{L}_{\mathrm{meta}}}
            {\left\|\nabla_{\widetilde{\theta}_t}
                    \mathcal{L}_{\mathrm{meta}}\right\|_2}.
  \label{eq:meta_loss}
\end{align}
The map $\mathsf{A}_t$ mirrors the persistent optimizer's learning rate,
moments, step count, bias correction, epsilon, and decoupled weight decay. It
does not modify the actual optimizer state. The detached result is therefore a
candidate student after one Fixed-weight KD step, not a persistent update. The
meta gradient does not pass through the AdamW transition. DIW evaluates the
candidate in inference mode on a right-arm mini-batch
$\mathcal{B}_m\subset\dmeta$ using the fitting-set class weights. The normalized
meta gradient defines the reference direction used by the probes. If its
norm is below $\delta=10^{-12}$, DIW sets both gate vectors to zero for the
current step.

\paragraph{Component-wise directional probes.}
The meta-feedback direction is evaluated after the candidate update. DIW then
asks how each sample's logit and feature losses change along that direction at
the persistent student. With $\xi=0.01$,
$\epsilon_{32}=2^{-23}$, $k\in\{L,F\}$, and
$\ell_{i,k}^{\pm}=\ell_i^k(\theta_t^{\pm})$, DIW computes
\begin{align}
  \theta_t^{\pm}
    &=\theta_t\pm\xi\widehat{\mathbf{g}}_m,
  \notag\\
  I_{i,k}
    &=\frac{\ell_{i,k}^{+}-\ell_{i,k}^{-}}{2\xi},
  \notag\\
  \tau_{i,k}
    &=64\epsilon_{32}
      \frac{|\ell_{i,k}^{+}|+|\ell_{i,k}^{-}|}{2\xi}.
  \label{eq:influence}
\end{align}
The positive and negative training-mode probes replay identical Python,
NumPy, PyTorch, and CUDA random-number states. They also start from independent
clones of the same persistent batch-normalization buffers. Thus dropout noise
does not determine the central difference. Probe losses are produced in FP32;
$I_{i,k}$ and its round-off resolution threshold $\tau_{i,k}$ are evaluated
in float64. To first order, $I_{i,k}>0$ means that decreasing component $k$
for sample $i$ agrees locally with decreasing the current meta objective. The
influence is a step-specific alignment measure, not an intrinsic property of
the sample or teacher target.

\paragraph{Positive evidence and the persistent update.}
For each distillation component, DIW keeps only positive influence that exceeds
the numerical resolution threshold. It then maximum-normalizes this evidence
within the current fitting mini-batch:
\begin{equation}
  \begin{aligned}
    e_{i,k}&=\max\{0,I_{i,k}-\tau_{i,k}\},
    \\[3pt]
    w_{i,k}&=
    \begin{cases}
      e_{i,k}\big/\displaystyle\max_{j\in\mathcal{B}_f}e_{j,k},
        & \displaystyle\max_{j\in\mathcal{B}_f}e_{j,k}>0,\\[2pt]
      0, & \text{otherwise}.
    \end{cases}
  \end{aligned}
  \label{eq:gate}
\end{equation}
The numerical result is clamped to $[0,1]$. The gates are normalized weights,
not probabilities. A zero gate means that the current probe found no resolved
positive evidence for that sample--component pair. It does not label the
sample or target as permanently harmful.

After restoring the pre-probe random-number state, DIW performs the persistent
training-mode forward pass at $\theta_t$ and minimizes
\begin{equation}
  \mathcal{L}_{\mathrm{DIW}}(\theta_t;\mathcal{B}_f)
  =\frac{1}{B}\sum_{i\in\mathcal{B}_f}
   \left[
     \ell_i^{\mathrm{CE}}(\theta_t)
     +\rho_e\!\sum_{k\in\{L,F\}}
       \lambda_k\stopgrad(w_{i,k})\ell_i^k(\theta_t)
   \right].
  \label{eq:diw_loss}
\end{equation}
The effective KD coefficient is $\lambda_k\rho_e w_{i,k}$. Since
$w_{i,k}\in[0,1]$, each DIW-weighted term is bounded by its Fixed-weight KD
counterpart and may be retained, attenuated, or suppressed.
Algorithm~\ref{alg:diw} summarizes the complete update order.

\begin{algorithm}[t]
  \caption{One DIW update after the supervised warm-up.}
  \label{alg:diw}
  \small
  \begin{algorithmic}[1]
    \Require current student parameters and buffers $(\theta_t,\mathbf{b}_t)$,
      AdamW state $\mathbf{s}_t$, fitting batch $\mathcal{B}_f$, matched-size
      meta batch $\mathcal{B}_m$, and cached teacher targets
    \Ensure updated persistent student and optimizer/scheduler states
    \State $\mathsf{r}\gets\Call{SnapshotRNG}{}$;
      $\widetilde{\mathbf{b}}\gets\Call{Clone}{\mathbf{b}_t}$
    \State Evaluate $\mathcal{L}_{\mathrm{fixed}}$ functionally on
      $\mathcal{B}_f$ in training mode using $(\theta_t,\widetilde{\mathbf{b}})$
    \State $\mathbf{q}_t\gets
      \operatorname{clip}_{C}(\nabla_{\theta_t}\mathcal{L}_{\mathrm{fixed}})$
    \State $\widetilde{\theta}_t\gets
      \stopgrad\!\left[\mathsf{A}_t(\theta_t,\mathbf{q}_t;\mathbf{s}_t)\right]$
      \Comment{$\mathbf{s}_t$ is unchanged}
    \State $\mathbf{g}_m\gets\nabla_{\widetilde{\theta}_t}
      \mathcal{L}_{\mathrm{meta}}(\widetilde{\theta}_t;\mathcal{B}_m)$
      using $(\widetilde{\theta}_t,\widetilde{\mathbf{b}})$
      \Comment{inference mode}
    \If{$\|\mathbf{g}_m\|_2<\delta$}
      \State $w_{i,L}\gets0$ and $w_{i,F}\gets0$
        for every $i\in\mathcal{B}_f$
    \Else
      \State $\widehat{\mathbf{g}}_m\gets
        \mathbf{g}_m/\|\mathbf{g}_m\|_2$
      \For{$\sigma\in\{+1,-1\}$}
        \State \Call{RestoreRNG}{$\mathsf{r}$};
          $\mathbf{b}^{\sigma}\gets\Call{Clone}{\mathbf{b}_t}$
        \State Evaluate $\{\ell_{i,L}^{\sigma},\ell_{i,F}^{\sigma}\}_{i\in\mathcal{B}_f}$
          in training mode at $\theta_t+\sigma\xi\widehat{\mathbf{g}}_m$
          using $\mathbf{b}^{\sigma}$
      \EndFor
      \For{$k\in\{L,F\}$}
        \State Compute $I_{i,k}$ and $\tau_{i,k}$ for all
          $i\in\mathcal{B}_f$ using equation~\eqref{eq:influence}
        \State Map positive evidence to component-specific gates
          $w_{i,k}$ using equation~\eqref{eq:gate}
      \EndFor
    \EndIf
    \State \Call{RestoreRNG}{$\mathsf{r}$}; evaluate
      $\mathcal{L}_{\mathrm{DIW}}$ on the persistent student in training mode
    \State Update the persistent student and AdamW state with
      $\operatorname{clip}_{C}(\nabla_{\theta_t}\mathcal{L}_{\mathrm{DIW}})$;
      advance the learning-rate scheduler
  \end{algorithmic}
\end{algorithm}

\subsection{Shared training state and inference contract}
\label{sec:deployment_contract}

All three right-arm routes use the same architecture, seed, and data order. They follow an identical supervised trajectory through epoch 5, including the model, optimizer, scheduler, batch-normalization, and random-number states. Fixed-weight KD and DIW become active from epoch 6 and share the teacher, component losses, base coefficients, and ramp.

After training, only the right-arm student is retained. The four-IMU teacher, $\dmeta$, look-ahead state, probes, and non-right-arm streams are absent from inference. Supervised, Fixed-weight KD, and DIW therefore use the same right-arm input tensor and 80,915-parameter forward graph; DIW changes only the training procedure.

\FloatBarrier
\section{Experimental Setup}
\label{sec:experiments}

\subsection{Dataset and preprocessing}

We use the WEAR outdoor-sports activity-recognition dataset~\cite{bock2024wear}. The evaluated cohort contains 22 participants and 19 labels: 18 exercise activities and the null class. We use synchronized acceleration from IMUs at the right arm, right leg, left leg, and left arm; video is not used. We treat the released acceleration streams as 50-Hz signals. We form non-overlapping 1.0-s windows of 50 samples (stride 50), then convert each position to the seven channels defined in Section~\ref{sec:networks}. Each window is assigned its majority sample-level label, with the center sample breaking ties.

The initial synchronized cohort contains 69,326 physical windows. Before constructing fold-specific subsets, we remove every window with a non-finite value at any required position. This common filter removes 1,028 windows from one participant and leaves 68,298 complete windows under the fixed outer-fold assignments. All routes use the same window keys, labels, participant identifiers, and assignments. No data augmentation is applied.

\subsection{Subject-disjoint protocol}

We use five-fold StratifiedGroupKFold with seed 42 and participant identity as the grouping variable~\cite{pedregosa2011scikit}. The five outer test folds contain 10,321, 16,636, 12,104, 16,714, and 12,523 windows. Within each outer-training split, three pre-specified participants form $\dmeta$, and the remaining participants form $\dfit$. For folds 0--4, the meta-participant sets are respectively \{sbj\_12, sbj\_17, sbj\_6\}, \{sbj\_11, sbj\_16, sbj\_9\}, \{sbj\_11, sbj\_16, sbj\_7\}, \{sbj\_11, sbj\_16, sbj\_5\}, and \{sbj\_12, sbj\_17, sbj\_9\}. Across folds, $\dfit$ contains 43,702--49,113 windows and $\dmeta$ contains 7,882--8,864 windows. Every partition retains all 19 labels.

Each DIW meta mini-batch has the same size as the corresponding fitting mini-batch. The sampler allocates nearly equal quotas to the three meta participants through deterministic participant-wise cycles, followed by a deterministic permutation. The meta set is reserved for DIW, which accesses only its right-arm windows and labels through equation~\eqref{eq:meta_loss}; Supervised and Fixed-weight KD train only on $\dfit$. We evaluate the final epoch in each outer fold without checkpoint selection on $\dtest$.

\subsection{Optimization and comparison routes}

Students are trained for 32 epochs with AdamW~\cite{loshchilov2019adamw}. We use $(\beta_1,\beta_2)=(0.9,0.999)$, $\epsilon_{\mathrm{opt}}=10^{-8}$, an initial learning rate of $10^{-3}$, and weight decay of $10^{-4}$. The learning rate follows cosine annealing to $10^{-5}$ over all optimizer steps, and the scheduler advances after every mini-batch. Training and evaluation batch sizes are 512 and 4,096, respectively; the global gradient norm is clipped at 5.0. Runs use deterministic FP32 computation in PyTorch 2.5.1 with CUDA 12.4; automatic mixed precision and TF32 are disabled. Each teacher stage uses the same core optimizer settings and runs for 32 epochs, as described in Section~\ref{sec:networks}.

We compare three deployable routes:
\begin{enumerate}[leftmargin=1.55em,itemsep=0.2em,topsep=0.3em]
  \item \textbf{Supervised}: the right-arm student optimized with equation~\eqref{eq:hard_loss} alone;
  \item \textbf{Fixed-weight KD}: the same student optimized with equation~\eqref{eq:fixed_kd}; and
  \item \textbf{DIW}: the same student and distillation components optimized with equation~\eqref{eq:diw_loss}.
\end{enumerate}
The four-IMU teacher is reported separately as a non-deployable reference because its predictions require all four IMUs and its capacity and fusion graph differ from those of the students.

\subsection{Metrics and uncertainty}

The primary metric is pooled OOF macro-F1, which weights each label equally~\cite{sokolova2009measures}. We concatenate predictions from the five mutually exclusive outer test folds and recompute F1 over the fixed 19-label space. We also report the unweighted mean and sample standard deviation of the five fold-level macro-F1 scores. Absolute differences in macro-F1 are described as percentage points.

For aggregate and per-label intervals, we resample the 22 participants as clusters 10,000 times with seed 42. We recompute each statistic on every resampled cohort. Table~\ref{tab:main_results} reports percentile 95\% intervals for pooled macro-F1, and Figure~\ref{fig:breadth} applies the same procedure to per-label F1. For each per-participant interval, windows are instead resampled within that participant's observed labels. These intervals describe uncertainty in each displayed estimate; we report route differences descriptively rather than as paired significance tests~\cite{efron1993bootstrap}.

\subsection{Mechanism diagnostic and ablations}

The mechanism diagnostic records normalized gates $w_{i,L}$ and $w_{i,F}$ for all 49,113 Fold-0 fitting samples at epochs 6, 8, 16, and 32. Exact zeros are retained, and a gate is positive when $w_{i,k}>0$. For each activity class and component, the class-conditional mean includes every fitting sample of that class, including samples with zero gates. We report Fold 0 as a detailed case study of training behavior, not as a separate accuracy estimate or a cross-fold mechanism claim.

The ablation study retains the same teacher, student, component losses, base coefficients, and optimization protocol. \emph{Mean-matched KD} assigns every fitting sample the fold-, epoch-, and component-specific mean of $\lambda_k\rho_e w_{i,k}$ recorded from the corresponding Full DIW run. It therefore matches the mean effective component coefficient, but not the weighted loss or gradient, without sample-wise assignment or online assessment. \emph{Static gate map} estimates the sample-wise gates once when DIW begins at epoch 6 and reuses the resulting map thereafter. \emph{No look-ahead} obtains the meta-loss direction from the current student rather than the detached look-ahead state. \emph{Batch-shared gates} recomputes the gates online but replaces the sample-wise values for each component by their detached mini-batch mean. \emph{Full DIW} retains step-wise re-estimation, the look-ahead state, and sample-wise gates.

\section{Results}
\label{sec:results}

\subsection{Selective distillation improves the right-arm student}

DIW achieves the strongest deployable right-arm result under the common protocol (Table~\ref{tab:main_results}). Supervised training reaches a pooled OOF macro-F1 of 0.561820. Fixed-weight KD raises it by 0.98 percentage points to 0.571623. DIW reaches 0.638451, gains of 7.66 points over Supervised and 6.68 points over Fixed-weight KD. The five-fold mean follows the same ordering. Both KD routes use the same teacher, targets, base coefficients, and ramp. DIW additionally uses labels from fold-internal meta participants to estimate its gates.

\begin{table}[H]
  \caption{Subject-disjoint performance with right-arm-only student inference.}
  \label{tab:main_results}
  \centering
  \small
  \begin{tabular*}{\linewidth}{@{\extracolsep{\fill}}lllccc@{}}
    \toprule
    Method & Train privilege & Test input & OOF F1 & Fold F1 & 95\% CI \\
    \midrule
    Supervised          & --            & RA    & 56.18 & $56.10\!\pm\!1.64$ & 52.00--60.44 \\
    Fixed-weight KD & Four-IMU teacher & RA       & 57.16 & $57.17\!\pm\!2.66$ & 52.89--61.49 \\
    \textbf{DIW}    & Four-IMU teacher & RA       & \textbf{63.85} & $\mathbf{63.97\!\pm\!3.27}$ & \textbf{59.54--68.06} \\
    \addlinespace[2pt]
    \midrule
    Four-IMU teacher    & --               & Four IMUs & 70.48 & $70.60\!\pm\!2.65$ & 66.39--74.49 \\
    \bottomrule
  \end{tabular*}
  \vspace{2pt}
  \begin{minipage}{\linewidth}
    \footnotesize
    F1 values are percentages. OOF F1 is pooled over 68,298 windows; fold F1 is mean $\pm$ sample SD over five folds. CIs use 10,000 participant-cluster bootstrap resamples. RA denotes right-arm IMU input. The teacher is a non-deployable reference.
  \end{minipage}
\end{table}

\subsection{Gains extend across labels and held-out participants}

\begin{figure}[H]
  \centering
  \includegraphics[width=0.94\linewidth]{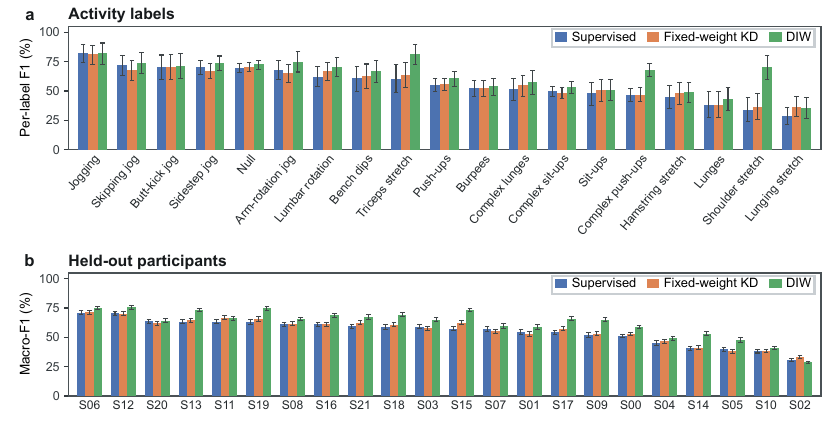}
  \caption{Performance across labels and held-out participants. (a) OOF per-label F1 for 18 activities and the null class; (b) per-participant OOF macro-F1 for 22 participants. Groups are ordered by Supervised performance. Error bars show 95\% participant-cluster bootstrap intervals in (a) and descriptive within-participant resampling intervals in (b).}
  \label{fig:breadth}
\end{figure}

Figure~\ref{fig:breadth} resolves the pooled result across all 19 labels and 22 held-out participants on the common OOF cohort.

The DIW gain is broad but not universal across the evaluation cohort. Relative to Supervised, Fixed-weight KD improves 12 of 19 labels and 15 of 22 held-out participants. DIW improves 18 labels and 21 participants. It also exceeds Fixed-weight KD for 18 labels and 20 participants.

All three students use identical inputs, parameter counts, and forward graphs at inference, holding deployed capacity fixed across the comparison. The four-IMU teacher reaches 0.704782, but its richer input and larger fusion model make it a contextual reference rather than a directly comparable student route.

The largest DIW gains over Supervised occur for Shoulder stretch (+37.00 percentage points), Complex push-ups (+21.39 points), and Triceps stretch (+21.23 points). Jogging changes by $-0.03$ points, and Lunging stretch is $0.97$ points below Fixed-weight KD. At participant level, S02 is 2.28 points below Supervised and also below Fixed-weight KD. S11 is 0.46 points below Fixed-weight KD. These exceptions show that the aggregate improvement does not remove all label- or participant-specific failures.

\subsection{Gate selection varies by component, stage, and activity class}

\begin{figure}[H]
  \centering
  \includegraphics[width=\linewidth]{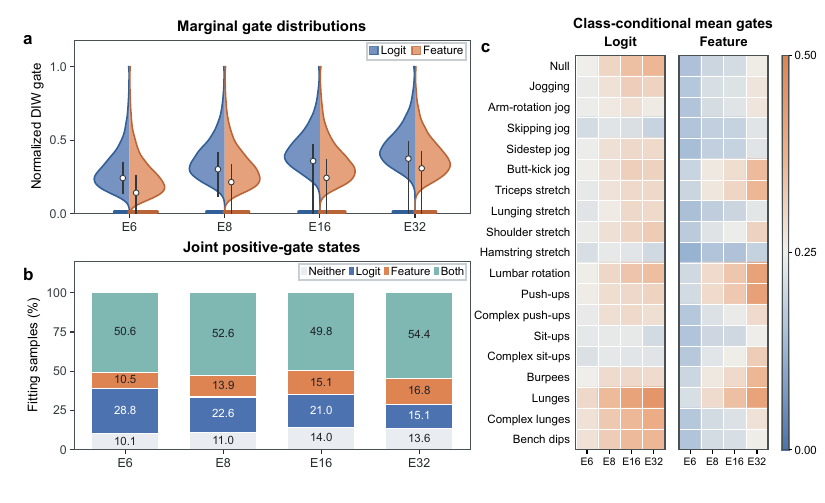}
  \caption{Fold-0 DIW gate dynamics. (a) Normalized logit and feature gates at epochs 6, 8, 16, and 32; basal segments denote zero mass, and points with bars denote the all-sample median and interquartile range. (b) Fractions of fitting samples in the four joint positive-gate states. (c) Class-conditional mean logit and feature gates, including zeros. The shared 0.00--0.50 color scale reports unitless mean gate values. All panels use the 49,113 Fold-0 fitting samples.}
  \label{fig:gate_diagnostics}
\end{figure}

Figure~\ref{fig:gate_diagnostics} resolves Fold-0 gate behavior across marginal, joint-state, and class-conditional views.

Panel (a) reports gate frequency and magnitude separately. The positive logit-gate rate decreases from 79.4\% at epoch 6 to 69.5\% at epoch 32. Over the same period, its all-sample median rises from 0.242 to 0.374. Positive logit gates therefore become less frequent while the overall median increases. For feature gates, the positive rate rises from 61.1\% to 71.3\%, and the median rises from 0.142 to 0.308. The two components follow different training trajectories.

Panel (b) shows that feature gates do not simply replace logit gates. The both-positive state remains the largest throughout training (49.8--54.4\%). Logit-only incidence falls from 28.8\% to 15.1\%, whereas feature-only incidence rises from 10.5\% to 16.8\%. Panel (c) further resolves these changes by activity. The mean normalized logit gate rises from 0.247 to 0.322, and the feature mean rises from 0.164 to 0.282. Logit means are higher for all 19 classes at epoch 6. By epoch 32, feature means are higher for 10 classes. These class-specific trajectories describe Fold-0 gate behavior and do not by themselves explain class-level accuracy gains.

\FloatBarrier

\subsection{Full DIW exceeds coefficient- and structure-reduced controls}

The ablations distinguish adaptive selection from a reduction in average KD strength (Table~\ref{tab:ablations}). Mean-matched KD reaches 0.599342, improving on Fixed-weight KD but remaining 3.91 percentage points below Full DIW. The Static gate map is the strongest reduced variant at 0.625185, 1.33 points below Full DIW. No look-ahead and Batch-shared gates reach 0.616044 and 0.606234, respectively.

\begin{table}[H]
  \caption{Ablation of dynamic influence weighting.}
  \label{tab:ablations}
  \centering
  \small
  \begin{tabular*}{\linewidth}{@{\extracolsep{\fill}}lcccccc@{}}
    \toprule
    Variant & Online & Look-ahead & Sample-wise & OOF F1 & Fold F1 & $\Delta$ \\
    \midrule
    Mean-matched KD  & --         & --         & --          & 59.93 & $60.00\!\pm\!1.92$ & $-3.91$ \\
    Static gate map  & --         & \checkmark & \checkmark  & 62.52 & $62.53\!\pm\!2.48$ & $-1.33$ \\
    No look-ahead    & \checkmark & --         & \checkmark  & 61.60 & $61.64\!\pm\!2.62$ & $-2.24$ \\
    Batch-shared gates & \checkmark & \checkmark & --        & 60.62 & $60.56\!\pm\!1.97$ & $-3.22$ \\
    \textbf{Full DIW} & \checkmark & \checkmark & \checkmark & \textbf{63.85} & $\mathbf{63.97\!\pm\!3.27}$ & -- \\
    \bottomrule
  \end{tabular*}
  \vspace{2pt}
  \begin{minipage}{\linewidth}
    \footnotesize
    F1 and $\Delta$ values are percentages; $\Delta$ is relative to Full DIW. ``Online'' denotes step-wise gate re-estimation, and ``Sample-wise'' denotes within-batch sample-specific gates. All variants use the same four-IMU teacher, right-arm student, distillation components, and optimizer; student inference remains right-arm-only.
  \end{minipage}
\end{table}

Mean-matched KD recovers part of the improvement, so reducing average KD strength is useful but does not reproduce Full DIW. The Static gate map is the strongest reduced variant. Its 1.33-point gap to Full DIW is consistent with a benefit from recalculating the gates as the student changes. The lower \emph{No look-ahead} result is consistent with value from evaluating guidance after the candidate optimizer step. The Batch-shared result is consistent with information being lost when one component weight is assigned to an entire mini-batch. These are targeted comparisons of complete variants, not a factorial estimate of each component's independent causal effect.

\section{Discussion}
\label{sec:discussion}

The experiments show that a stronger multi-position teacher does not automatically produce a much stronger single-position student. The teacher exceeds the supervised student by 14.30 percentage points, but Fixed-weight KD recovers only 0.98 points. This gap is plausible because the teacher observes motion that is absent from the right-arm input. A target produced by an accurate four-IMU teacher may still fail to provide a useful update for the restricted student. DIW's larger gain is consistent with controlling each target according to the student's current response instead of applying it uniformly.

The gate diagnostics and ablations support this interpretation in complementary ways. Logit and feature gates follow different trajectories, so one shared schedule cannot express their observed behavior. The both-positive state remains common, which suggests that the two knowledge sources often remain complementary rather than mutually exclusive. Mean-matched KD does not reproduce Full DIW, so the result is not explained by lower average KD strength alone. Static gates retain much of the gain, while online gates perform better as the student changes. These observations support sample- and component-specific selection, but they do not prove that a particular class-level gate pattern caused a particular accuracy gain.

This result complements prior work that uses additional sensors during training while retaining one sensor at inference~\cite{lago2021additional,rey2022learning,nguyen2024virtual}. It emphasizes that the way teacher guidance enters the student update matters when teacher and student observe different body locations. DIW leaves the deployed input and model unchanged, but it increases training cost. Every post-warm-up step requires a candidate AdamW update, a meta-batch evaluation, and two finite-difference probes. The method is therefore most relevant when richer sensing and additional computation are acceptable during model development. Hardware-normalized training time, memory use, and on-device performance remain to be measured.

Several limitations constrain the interpretation. The study uses one dataset, one retained sensor location, one random seed, and one teacher--student family. Only DIW uses labels from the fold-internal meta participants. Its gain over Supervised and Fixed-weight KD may therefore reflect both the gating procedure and access to additional labeled feedback. A control with equivalent access to meta labels is needed to separate these effects. Figure~\ref{fig:gate_diagnostics} describes only Fold 0, and the ablations compare targeted variants rather than a full factorial design. Further evaluation should include additional seeds, datasets, retained locations, training-cost measurements, and on-device profiling.

\section{Conclusion}
\label{sec:conclusion}

We investigated whether multi-position IMU signals available during training can improve a classifier deployed with only a right-arm IMU. DIW assigns separate sample-wise gates to logit and feature targets using feedback from a one-step candidate update. Under subject-disjoint five-fold evaluation on WEAR, it raises pooled OOF macro-F1 from 0.561820 to 0.638451. The deployed model remains the same 80,915-parameter right-arm student. The ablations show that lowering average KD strength alone does not reproduce the Full DIW result. These findings support selective privileged distillation under the tested protocol. Equivalent meta-label access and broader replication are needed to isolate the source of the gain and establish its generality.

\section*{Data Availability}

The WEAR dataset is available from the official project website and is described by Bock \emph{et al.}~\cite{bock2024wear}.

\section*{Code Availability}

A PyTorch reference implementation of the right-arm student, four-IMU teacher, three training objectives, and DIW meta-probe and component-wise gate mapping is publicly available.\par\smallskip
\noindent Repository: \href{https://github.com/1304126986/wear-right-arm-privileged-distillation}{\nolinkurl{github.com/1304126986/wear-right-arm-privileged-distillation}}.

\bibliographystyle{unsrtnat}
\renewcommand{\bibfont}{\small\setlength{\baselineskip}{9.2pt}}
\setlength{\bibsep}{0pt}
\bibliography{references}

\end{document}